\pdfoutput=1

\documentclass[11pt]{article}

\usepackage{EMNLP2023}
\usepackage{pifont}
\usepackage{times}
\usepackage{latexsym}
\usepackage{fontawesome}
\usepackage[T1]{fontenc}
\usepackage{marvosym}
\usepackage{hyperref}
\usepackage{enumitem}
\usepackage{amsmath}
\usepackage{algorithmic}
\usepackage{algorithm}
\usepackage{tcolorbox}
\usepackage{colortbl}
\usepackage{booktabs}
\usepackage{multirow}
\usepackage{makecell}
\usepackage{balance}
\usepackage{soul, color, xcolor}

\usepackage[utf8]{inputenc}

\usepackage{microtype}

\usepackage{inconsolata}
\usepackage{graphicx}
\usepackage{marvosym}
\usepackage{amssymb}

\newcommand{\nosection}[1]{\vspace{5pt}\noindent\textbf{#1.}}

\title{\textsc{Medico}: Towards Hallucination Detection and Correction with Multi-source Evidence Fusion}

\author{Xinping Zhao$^{1}$,  Jindi Yu$^{1}$, Zhenyu Liu$^{1}$, Jifang Wang$^{1}$, \\  \textbf{Dongfang Li$^{1}$, Yibin Chen$^{2}$, Baotian Hu$^{1}$\textsuperscript{\Letter}\thanks{\textsuperscript{\Letter}Corresponding author.},  Min Zhang$^{1}$} \\ 
        $^{1}$Harbin Institute of Technology (Shenzhen), Shenzhen, China, \\ $^{2}$Huawei Cloud, Huawei Technologies Ltd. \\
        \texttt{\{zhaoxinping, 22S051013, 190110924, 23S151116\}@stu.hit.edu.cn}, \\ chenyibin4@huawei.com, \texttt{\{lidongfang, hubaotian, zhangmin2021\}@hit.edu.cn}}

\makeatletter
\def\thanks#1{\protected@xdef\@thanks{\@thanks
        \protect\footnotetext{#1}}}
\makeatother

\begin{document}
\maketitle
\begin{abstract}
As we all know, hallucinations prevail in Large Language Models (LLMs), where the generated content is coherent but factually incorrect, which inflicts a heavy blow on the widespread application of LLMs. 
Previous studies have shown that LLMs could confidently state non-existent facts rather than answering ``I don't know''. Therefore, it is necessary to resort to external knowledge to detect and correct the hallucinated content. Since manual detection and correction of factual errors is labor-intensive, developing an automatic end-to-end hallucination-checking approach is indeed a needful thing.
To this end, we present \textsc{Medico}, a \underline{M}ulti-source \underline{e}vidence fusion enhanced hallucination \underline{d}etect\underline{i}on and \underline{co}rrection framework.
It fuses diverse evidence from multiple sources, detects whether the generated content contains factual errors, provides the rationale behind the judgment, and iteratively revises the hallucinated content.
Experimental results on evidence retrieval (0.964 HR@5, 0.908 MRR@5),  hallucination detection (0.927-0.951 F1), and hallucination correction (0.973-0.979 approval rate) manifest the great potential of \textsc{Medico}. A video demo of \textsc{Medico} can be found at \url{https://youtu.be/RtsO6CSesBI}.
\end{abstract}

\section{Introduction}
Large Language Models (LLMs) have attracted significant interest from academia and industry. Major tech companies have introduced solutions like OpenAI's GPT-4 \cite{DBLP:journals/corr/abs-2303-08774}, Google's Gemini \cite{DBLP:journals/corr/abs-2403-05530}, and Alibaba's Qwen \cite{qwen2,DBLP:journals/corr/abs-2309-16609}. LLMs have shown impressive performance in understanding and generating language. 
However, their complex structures, vast parameters, and opaque generation processes make it difficult to ensure the accuracy of the generated content, known as hallucination\footnote{Hallucination can be broadly categorized into \textit{Factuality Hallucination} and \textit{Faithfulness Hallucination}, referring to Section \ref{sec:hallu_in_llms} for more details.
This work mainly focuses on \textit{Factuality Hallucination}.} \cite{DBLP:journals/corr/abs-2311-05232,DBLP:conf/emnlp/MinKLLYKIZH23,DBLP:journals/corr/abs-2402-09733}, posing potential risks for widespread practical application. Hence, developing a robust hallucination-checking approach to verify LLMs' generated content has become one of the crucial challenges that need to be addressed urgently \cite{DBLP:journals/corr/abs-2402-02420,DBLP:journals/corr/abs-2311-09000}. 
\begin{figure}[t]
    \centering
    \includegraphics[width=1.\linewidth]{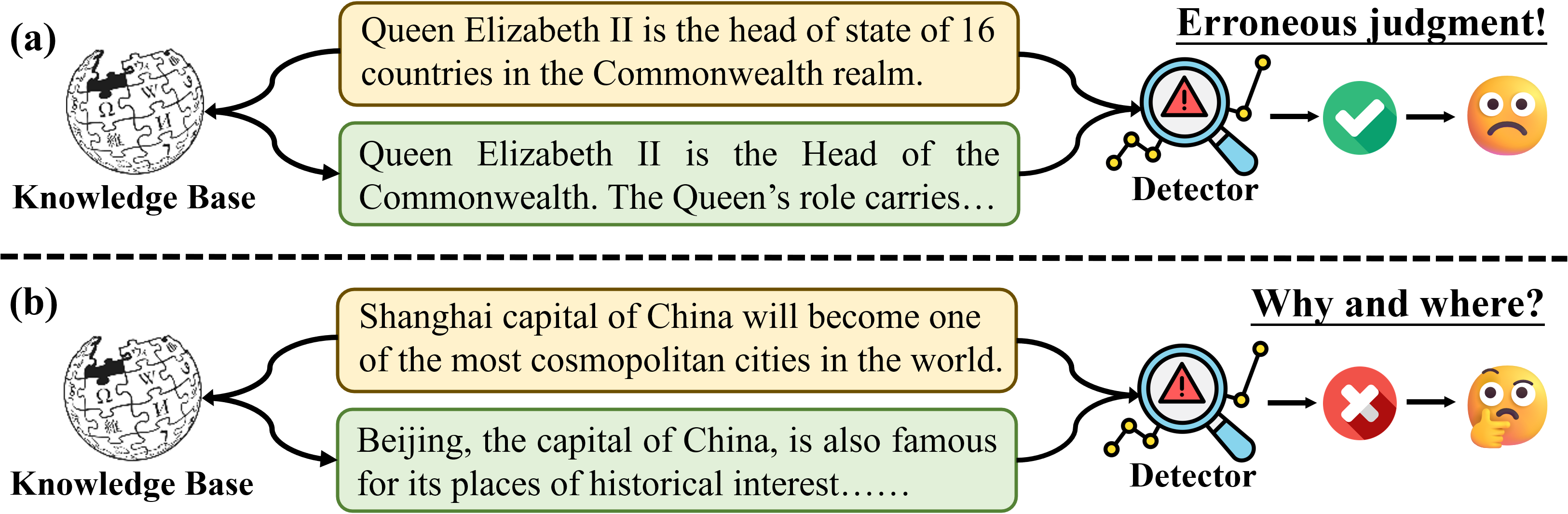}
    \caption{Motivation example. The generated content and retrieved evidence are marked in \textbf{\textcolor[RGB]{194,144,0}{yellow}} and \textbf{\textcolor[RGB]{0,160,0}{green}}, respectively. (a) shows the situation of acquiring evidence in a single way and making an erroneous judgment due to outdated evidence. (b) shows the situation, where users are only provided with a veracity label, confusing users about why and where the content is incorrect.} 
    \label{fig:example}
    \vspace{-0.2cm}
\end{figure}
Recently, an ever-growing body of studies and systems has been focused on verifying LLMs' generated content in terms of hallucinations, such as FLEEK \cite{DBLP:conf/emnlp/BayatQHSBKWI023}, FactLLaMA \cite{DBLP:conf/apsipa/CheungL23}, and SAFE \cite{DBLP:journals/corr/abs-2403-18802}. They formulate hallucination-checking as the classification task, where the input consists of the evidence and generated content, and the output typically determines the veracity of the generated content into three categories, \textit{i.e.,} SUPPORTED, NOT SUPPORTED, and IRRELEVANT \cite{DBLP:conf/naacl/ThorneVCM18}. However, they commonly acquire evidence in a single way and may fall into the absence of useful evidence. In fact, the accuracy of the generated content involves many aspects, requiring informative evidence from diverse sources. 
Taking the generated content ``\textit{Queen Elizabeth \uppercase\expandafter{\romannumeral2} is the head of state of 16 countries in the Commonwealth realm.}'' for example, it might be classified as correct when only using evidence acquired from a non-real-time knowledge base, as shown in Figure \ref{fig:example}(a). On the other hand, they usually show users only the veracity label, while the rationale behind such a decision is missing. So these models lack explainability and still require arduous labor from users to manually check why and where the generated content is incorrect, which creates a poor user experience. We show this issue in Figure \ref{fig:example}(b). 
In this work, we propose \textsc{Medico} (\underline{M}ulti-source \underline{e}vidence fusion enhanced hallucination \underline{d}etect\underline{i}on and \underline{co}rrection), a hallucination-checking framework, which satisfies the three properties of being multi-faceted, model-agnostic, and explainable. 
Specifically, our framework acquires diverse evidence from multiple sources, including unstructured text, semi-structured knowledge base, as well as structured knowledge graphs. 
It reranks the evidence candidates and organically fuses them to obtain the fused evidence, which offers sufficient support evidence for the following detection.
Our framework then leverages the fused evidence to detect whether the generated content is correct or incorrect and also gives the rationale behind the decision.
If the classification result is incorrect, it will iteratively revise the hallucinations within the generated content according to the rationale.
Our main contributions can be summarized as follows: 
\begin{itemize}[leftmargin=*]
    \item To the best of our knowledge, the proposed \textsc{Medico} is the first hallucination detection and correction framework that performs multi-source evidence fusion, provides the rationale behind the decision,\, and\, corrects\, the hallucinated\, content.
    \item Our \textsc{Medico} is highly user-friendly and explainable, where users only need to provide the generated content and all data flow from evidence retrieval to decision-making\, could\, be\, traceable. 
    \item Our \textsc{Medico} is model-agnostic and can adopt any off-the-shelf LLMs to conduct evidence fusion and hallucination detection and correction. 
    \item We conduct extensive experiments on HaluEval \cite{DBLP:conf/emnlp/LiCZNW23}, whose results fully verify the effectiveness of the proposed \textsc{Medico} in terms of retrieval, detection, and correction performance.
\end{itemize}
\section{Methodology}
Figure \ref{fig:framework} presents the overall system framework of \textsc{Medico}. It mainly consists of three components: (1) Multi-source Evidence Fusion, which incorporates diverse evidence from multiple sources to provide sufficient support evidence for detection; 
(2) Hallucination Detection with Evidence, which leverages the fused evidence to check LLMs' generated content and gives the rationale behind the decision; 
(3) Hallucination Correction with Rationale, which iteratively revises the hallucinated content until the pre-defined threshold is reached or the revised content is approved by the detector.
\subsection{Multi-source Evidence Fusion}
\label{section:evidence_fusion}
Evidence can be retrieved from a closed knowledge base such as Wikipedia, using an open-domain search engine (\textit{e.g.,} Google and Bing), from a well-organized knowledge graph, or even user-uploaded files \cite{DBLP:journals/corr/abs-2311-09000}. 
Given that the accuracy of the generated content involves many aspects, it is necessary and valuable to acquire informative evidence from multiple sources. Afterward, we organically fuse them to eliminate varied writing styles since they come from diverse sources. 
Given a user query $q$ and the generated content $o$, we send them to our multi-source evidence fusion system, which is composed of evidence retrieval and fusion:
\begin{figure*}[t]
    \centering
    \includegraphics[width=1.\linewidth]{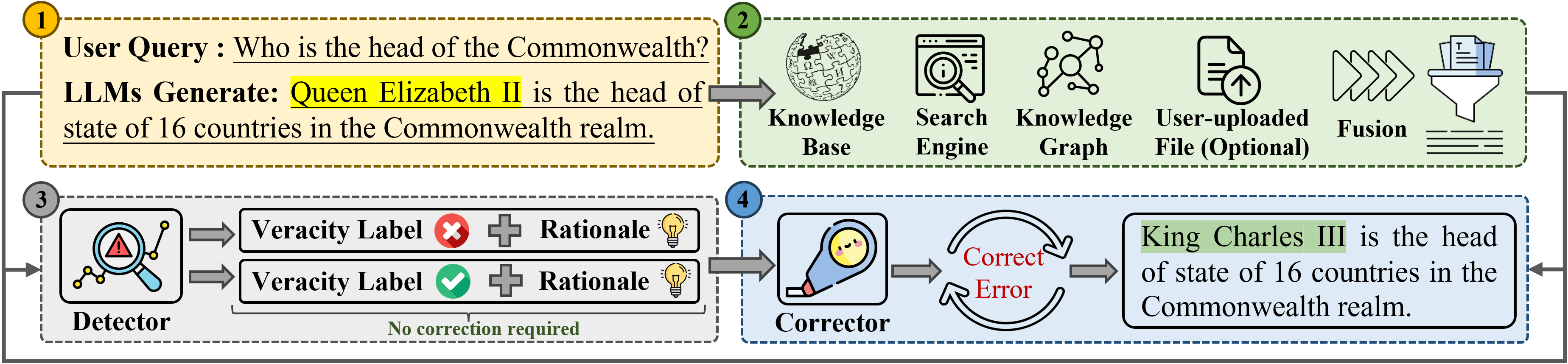}
    \caption{The overall system framework of \textsc{Medico}. The upper layer illustrates the working flow of multi-source evidence fusion while the bottom layer illustrates the working flow of hallucination detection as well as correction.} 
    \label{fig:framework}
    \vspace{-0.2cm}
\end{figure*}
\nosection{Evidence Retrieval} Here, we adopt diverse heterogeneous sources to retrieve evidence as informative as possible. Specifically, we build the retrieval system on four complementary\, sources\, as below:
\begin{itemize}[leftmargin=*]
    \item \textbf{Search Engine (Web).} We search top passages using Google Search API provided by Serper\footnote{\url{https://serper.dev/}}. Then, we recall the $n$ most relevant snippets ${E}^S = \{e^s_1, e^s_2, ..., e^s_{n}\}$ in API's Responses based on the user query $q$ and the generated content $o$.
    \item \textbf{Knowledge Base (KB).} We use the English Wikipedia\footnote{\url{https://huggingface.co/datasets/lsb/enwiki20230101}} from 01/01/2023 when the data annotation was completed, and we split each page into passages up to 256 tokens. Then, we retrieve the $m$ most\, relevant\, chunks\, ${E}^B = \{e^b_1, e^b_2, ..., e^b_{m}\}$.
    \item \textbf{Knowledge Graph (KG).} We utilize Wikidata5m \cite{DBLP:journals/tacl/WangGZZLLT21}, a million-scale knowledge graph, which consists of 4,594,485 entities, 822 relations and 20,624, 575 triples. Before retrieving, we first linearize triplets into passages using templates and then directly recall the $k$ most\, relevant\, ones ${E}^G = \{e^g_1, e^g_2, ..., e^g_{k}\}$.
    \item \textbf{User-uploaded File (UF).} In addition to the predetermined retrieval sources covered so far, users may need to use their customized ones, such as knowledge in a specialized field, when the user query is domain-specific. To this end, our framework further allows users to customize their desired retrieval sources. 
    Specifically, the system supports uploading files in four formats, \textit{i.e.,} TXT, DOCX, PDF, and MARKDOWN. Analogously, we retrieve the $j$ relevant chunks ${E}^U = \{e^u_1, e^u_2, ..., e^u_{j}\}$ from the user's uploaded files.
\end{itemize}
\nosection{Evidence Fusion} While multi-source retrieval can acquire abundant evidence, it can also draw a lot of noisy information, which may have a negative influence on the following hallucination detection. 
To address this issue, the evidence fusion aims for more accurate evidence by reranking the evidence set and fusing the top-ranked evidence. 
Specifically, we first combine all the evidence retrieved from diverse sources, which can be formulated as:
\begin{align}
\label{equ:combine}
  E &= \mathrm{Combine}(E^D|D\in\{S,B,G,U\})  \nonumber\\
   &= \{e_1, e_2, ..., e_{n+m+k+j}\},
\end{align}
where $D$ denotes the retrieval source, $E$ is the combined evidence set. Then, we re-rank the evidence set $E$ based on their relevance scores\footnote{We use bge-reranker-large \cite{bge_embedding} to measure the relevance score between the user query and the evidence.} with the user query. Afterward, we can get a newly ordered evidence set, which can be formulated as follows:
\begin{equation}
    \label{equ:rerank}
    \tilde{E} = \mathrm{Rerank}(q,o;E) = \{\tilde{e}_1, \tilde{e}_2, ..., \tilde{e}_{l}\},
\end{equation}
where $\tilde{e}_l$ denotes the evidence that has Top-$l$ relevance score among $E$, and $l\ll(n+m+k+j)$ denotes that the subset $\tilde{E}$ contains considerably fewer evidence than the original set $E$. Lastly, we fuse the reranked evidence set with concatenation or summarization, and we get the\, fused\, evidence:
\begin{equation}
    \label{equ:fuse}
    {E}^{F} = \mathrm{Fuse}(\tilde{E}),
\end{equation}
where we implement $\mathrm{Fuse}(\cdot)$ as concatenation or summarization. The former aims to preserve as much of the original evidence as possible. 
The latter aims for query-focused evidence summarization and eliminates the varied writing styles from diverse sources for better detection, where we find Llama3-8B-Instruct\, do\, well\, in\, summarizing\, $\tilde{E}$. 
\subsection{Hallucination Detection with Evidence}
Given the fused evidence ${E}^{F}$ and the generated content $o$, the detection task is to decide whether $o$ has factual errors conditioned on ${E}^{F}$, then provide the rationale behind this decision. Its working flow is shown in Figure \ref{fig:framework} lower left. Specifically, we implement hallucination detection in two manners:
\begin{figure*}[t]
    \centering
    \includegraphics[width=1.0\linewidth]{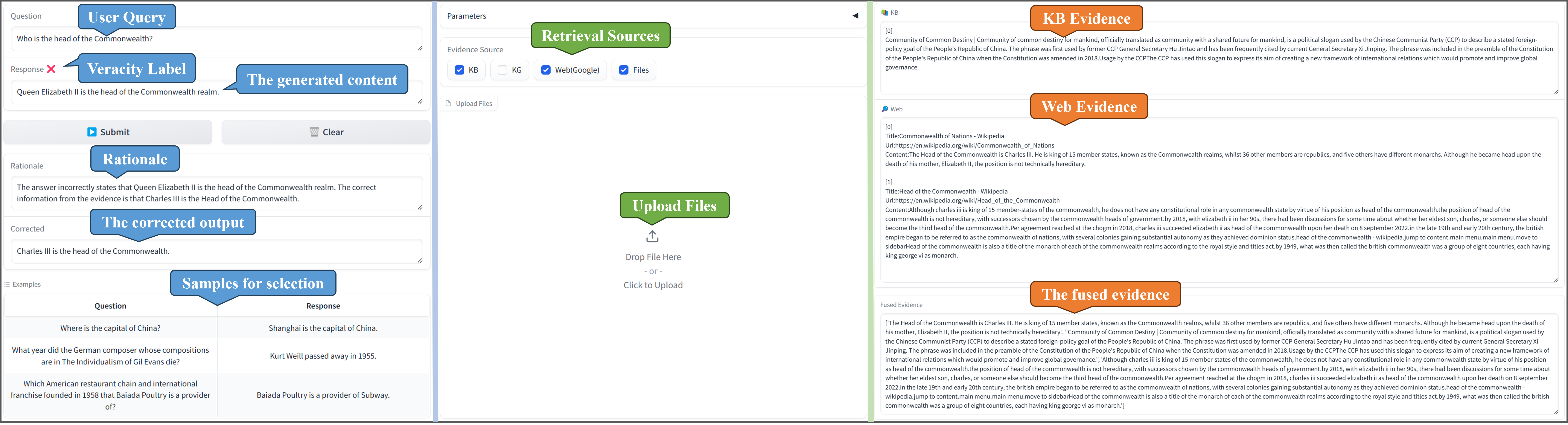}
    \caption{Screenshot of our hallucination detection and correction system \textsc{Medico}. The left shows the interface for entering the user query and the generated response. The middle shows the interface for selecting retrieval sources and uploading files. The right demonstrates the evidence retrieved from diverse\, sources\, and\, their\, fused\, evidence.} 
    \label{fig:user_interface}
    \vspace{-0.2cm}
\end{figure*}
\nosection{Detection with Fused Evidence} In this way, we directly prompt the detector, a designated LLM $\mathcal{M}_{d}$, to check whether the generated content conflicts with the fused evidence. If the output veracity label $v$ is False, it indicates that conflicts exist between ${E}^{F}$ and $o$. 
Afterward, we prompt $\mathcal{M}_{d}$ to generate the corresponding rationale $r$ that distinguishes the vital evidence from the fused evidence and explains how ${E}^{F}$ determines the veracity label $v$. Here, we employ in-context learning (ICL), a training-free technique \cite{dong2022survey}, which endows the detector model $\mathcal{M}_{d}$ with higher \quad \quad capacity to generate\, more\, reasonable\, rationale $r$.  
\nosection{Detection with Self-Consistency} To fully utilize the diversified evidence from multiple sources, we propose an ensemble method, which separately feeds the evidence derived from different sources into the detector $\mathcal{M}_{d}$ and learns to classify based on the likelihood collected from each source. Specifically, we first compute the\, likelihood\, as\, follows:
\begin{equation}
\label{equ:likehood}
    p(\mathrm{T}|q, o; E^*) = \frac{e^{\mathcal{M}_{d}(\mathrm{T}|q, o; E^*)/\tau}}{\sum_{v \in \{\mathrm{T}, \mathrm{F}\}} e^{\mathcal{M}_{d}(v|q, o; E^*)/\tau}},
\end{equation}
where $E^* \in \{E^S, E^B, E^G, E^U, E^F\}$; $\mathrm{T},\mathrm{F}$ denote True and False, respectively; $\tau$ is the temperature coefficient. Afterwards, we get $P = \{p^S, p^B, p^G, p^U, p^F\}$, where $P \in(0,1)^{5\times1}$ is the likelihood vector and each entry measures to what extent the generated content $o$ could be entailed by the evidence\footnote{We don't compute $p(\mathrm{F}|q, o; E^*)$ as it is complementary with $p(\mathrm{T}|q, o; E^*)$, where $p(\mathrm{T}|q, o; E^*)$+$p(\mathrm{F}|q, o; E^*)$=1.}. 
We build a binary classifier (\textit{i.e.,} Logistic Regression \cite{DBLP:books/wi/HosmerL00}) upon $P$ and use the binary cross-entropy (BCE) loss \cite{DBLP:journals/anor/BoerKMR05} to optimize the classifier:
\begin{equation}
\label{equ:bce}
    \mathcal{L}_{BCE}(y, \hat{y}) = y\log(\hat{y})+(1-y)\log(1-\hat{y}),
\end{equation}
where $y$ is the ground truth label, and $\hat{y}$ is the predicted probability of belonging to the positive class.
\subsection{Hallucination Correction with Rationale}
This module aims to correct the hallucinated parts in the generated content $o$ based on the rationale $r$, while the other parts remain unchanged. Its working flow is shown in Figure \ref{fig:framework} lower right. 
Inspired by \cite{DBLP:conf/acl/GaoDPCCFZLLJG23}, we adopt chain-of-thought (CoT), where we prompt the corrector model $\mathcal{M}_{c}$ to identify the hallucinated spans that need to be edited before correcting $o$. Then, we prompt $\mathcal{M}_{c}$ to revise these spans separately and output the corrected one $o^{\prime}$ that aims to agree with $r$. 
We perform multiple rounds of correction until the pre-defined threshold\footnote{Given the computational cost, we set the\, threshold\, as 5.} is reached or the detector $\mathcal{M}_{d}$ approves.
However, if not restrained, the corrector $\mathcal{M}_{c}$ may make superfluous modifications, such as reordering words, altering language style, and inserting unnecessary information \cite{DBLP:conf/acl/GaoDPCCFZLLJG23,DBLP:conf/acl/Thorne020}. 
To avoid excessive modifications on $o$, we first measure preservation using the variant of character-level Levenshtein edit distance \cite{DBLP:conf/acl/GaoDPCCFZLLJG23,levenshtein1966binary} as the metric, which can be formulated as follows:
\begin{equation}
\label{equ:prev}
    \mathrm{Prev}(o, o^{\prime}) = \mathrm{max}\left(1-\frac{\mathrm{Lev}(o, o^{\prime})}{\mathrm{Length}(o)}, 0\right),
\end{equation}
where $\mathrm{Lev}(\cdot)$ denotes the character-level Levenshtein edit distance function, $\mathrm{Prev}(\cdot)$ measures to what extent $o^{\prime}$ is consistent with $o$. If $\mathrm{Prev}(o, o^{\prime})$ equals 1.0, $o$ and $o^{\prime}$ are the same. On the other hand, if $\mathrm{Prev}(o, o^{\prime})$ equals 0.0, $o^{\prime}$ is totally different from $o$. 
During the iterative correction procedure, we reject those corrected outputs $o^{\prime}$, when $\mathrm{Prev}(o, o^{\prime})$ is\, less\, than\, $\delta$, a\, hyper-parameter\, to\, be\, adjusted.
\section{User Interface}
We build \textsc{Medico} using the Gradio package \cite{Abid2019GradioHS}, an easy-to-use WebUI development framework based on FastAPI and Svelte, which facilitates the deployment of machine learning apps. We can naturally divide the view of \textsc{Medico}'s system into two parts: (1) retrieval and fusion, and (2) detection and correction, as shown in Figure \ref{fig:user_interface}.
\nosection{Retrieval and Fusion View} To interact with \textsc{Medico}, users should first enter a query and the generated response into the corresponding box\footnote{As shown in Figure \ref{fig:user_interface}, we take the user query ``\textit{Who is the head of the Commonwealth?}'' for example. On the other hand, we take the generated content ``\textit{Queen Elizabeth \uppercase\expandafter{\romannumeral2} is the head of the Commonwealth realm.}'' as an example.}, or click one of the sample queries, as shown in the left side of Figure \ref{fig:user_interface}. 
Then, users can select the retrieval sources used, including Web, KB, and KG, as stated in \S\ref{section:evidence_fusion}, where users can also use their customized sources by uploading TXT, DOCX, PDF, and MARKDOWN from their local device (see the middle side of Figure \ref{fig:user_interface}). By the way, users can adjust the amount of evidence retrieved from each source and the amount of evidence to be used after the reranking, \textit{i.e.,} the hyper-parameter $l$. When the \textbf{Submit Button} is clicked, the evidence panel (see the right side of Figure \ref{fig:user_interface}) shows the evidence retrieved from each source and the fused evidence. 
\nosection{Detection and Correction View} In this view, \textsc{Medico} will request the hallucination detector model $\mathcal{M}_{d}$ to check whether the generated content $o$ contains factual errors conditioned on the fused evidence $E^F$ provided by the above. If there exist any factual errors, the detection panel will present the symbol of disapproval \textbf{\textcolor[RGB]{173,14,14}{\ding{56}}}, otherwise it will present the symbol of approval \textbf{\textcolor[RGB]{0,160,0}{\ding{52}}}.
Afterward, if \textsc{Medico} detects hallucinations, it will further request the hallucination corrector model $\mathcal{M}_{c}$ to correct them conditioned on the rationale or the fused evidence, where the rational $r$ and the corrected content $o^{\prime}$ will be displayed in the rationale panel as well as the correction\, panel,\, respectively.
\begin{table}[t]
  \renewcommand\arraystretch{1.25}
  \tabcolsep=0.148cm
  \footnotesize
  \centering
  \begin{tabular}{l|ccc|ccc}
    \toprule
    \multicolumn{1}{c|}{ \multirow{3}{*}[+0.ex]{\centering \bf \makecell[c]{Evidence \\ Sources}}} & 
    \multicolumn{6}{c}{ \multirow{1}{*}[+0.2ex]{\centering \bf Metrics}} \\
    \cline{2-7}
    
    & \multicolumn{3}{c|}{ \multirow{1}{*}[-0.275ex]{\centering \bf HR}} & \multicolumn{3}{c}{ \multirow{1}{*}[-0.275ex]{\centering \bf MRR}}\\
    \cline{2-7}
    
    & \multicolumn{1}{c}{ \multirow{1}{*}[-0.275ex]{\centering @1}} & \multicolumn{1}{c}{ \multirow{1}{*}[-0.275ex]{\centering @3}} & \multicolumn{1}{c|}{ \multirow{1}{*}[-0.275ex]{\centering @5}} & \multicolumn{1}{c}{ \multirow{1}{*}[-0.275ex]{\centering @1}} & \multicolumn{1}{c}{ \multirow{1}{*}[-0.275ex]{\centering @3}} & \multicolumn{1}{c}{ \multirow{1}{*}[-0.275ex]{\centering @5}} \\
    \cline{1-7}
   (A) Web & 0.458 & 0.589 & 0.637 & 0.458 & 0.518 &  0.529\\
   (B) KB & \underline{0.851} & \underline{0.903} &  \underline{0.909} & \underline{0.851} & \underline{0.876} & \underline{0.877} \\
   (C) KG & 0.639 & 0.675 &0.680 & 0.639 & 0.655 &  0.657\\
   (D) Fuse & \textbf{0.867} & \textbf{0.948} & \textbf{0.964} & \textbf{0.867} & \textbf{0.904} & \textbf{0.908} \\
    \bottomrule
  \end{tabular}
   \caption{Retrieval evaluation, where the best results are \textbf{boldfaced} and the second-best results are \underline{underlined}. The higher the metric score, the better the performance.}
  \label{tab:retrieval}
  \vspace{-0.2cm}
\end{table}
\section{Experiments}
In this section, we conduct extensive experiments on a hallucination evaluation benchmark, HaluEval, to answer the following Research Questions (\textbf{RQs}): 
\begin{itemize}[leftmargin=*]
    \item \textbf{RQ1:} Whether multi-source evidence retrieval can help improve the recall of golden evidence?
    \item \textbf{RQ2:} How does the fused evidence contribute to the hallucination detection performance in comparison with the evidence from a single source?
    \item \textbf{RQ3:} Can multi-turn editing and the generated rationale\, enhance\, the\, correction\, performance?
\end{itemize}
\subsection{Experimental Setup}
\nosection{Evaluation Data} We randomly sample 1000 <user query, right answer, hallucinated answer> triplet from HaluEval \cite{DBLP:conf/emnlp/LiCZNW23}, as evaluating the hit rate of evidence retrieval is labor-intensive. 
Then, we retrieve evidence from multiple sources (\textit{e.g.,} Web, KB, and KG) and perform evidence fusion, where we set $n,m,k,j$ as 5. We manually identify the golden evidence within the evidence set by checking whether it leads to the right answer. 
\nosection{Evaluation Metrics} For retrieval evaluation, we adopt two commonly used metrics: Hit Rate (HR) and Mean Reciprocal Rank (MRR). 
We also use the F1 score and approval rate as metrics to evaluate detection and correction performance, respectively.
\nosection{LLMs for Detection and Correction} We employ two different LLMs: Llama3-8B-Instruct\footnote{\url{https://github.com/meta-llama/llama3}} \cite{DBLP:journals/corr/abs-2407-21783} and Qwen2-7B-Instruct\footnote{\url{https://github.com/QwenLM/Qwen2}} \cite{qwen2}. We choose them as the hallucination detector $\mathcal{M}_d$ as well as hallucination corrector $\mathcal{M}_c$ because they are representative open-source LLMs\footnote{We use Llama3-8B and Qwen2-7B to represent Llama3-8B-Instruct and Qwen2-7B-Instruct, respectively, for brevity.}.
\begin{table}[t]
  \renewcommand\arraystretch{1.25}
  \tabcolsep=0.13cm
  \footnotesize
  \centering
  \begin{tabular}{l|ccc|ccc}
    \toprule
    \multicolumn{1}{c|}{ \multirow{3}{*}[+0.ex]{\centering \bf \makecell[c]{Evidence \\ Sources}}} & 
    \multicolumn{6}{c}{ \multirow{1}{*}[+0.2ex]{\centering \bf Detectors}} \\
    \cline{2-7}
    
    & \multicolumn{3}{c|}{ \multirow{1}{*}[-0.275ex]{\centering \bf Llama3-8B}} & \multicolumn{3}{c}{ \multirow{1}{*}[-0.275ex]{\centering \bf Qwen2-7B}}\\
    \cline{2-7}
    
    & \multicolumn{1}{c}{ \multirow{1}{*}[-0.275ex]{\centering Prec}} & \multicolumn{1}{c}{ \multirow{1}{*}[-0.275ex]{\centering Recall}} & \multicolumn{1}{c|}{ \multirow{1}{*}[-0.275ex]{\centering F1}} & \multicolumn{1}{c}{ \multirow{1}{*}[-0.275ex]{\centering Prec}} & \multicolumn{1}{c}{ \multirow{1}{*}[-0.275ex]{\centering Recall}} & \multicolumn{1}{c}{ \multirow{1}{*}[-0.275ex]{\centering F1}} \\
    \cline{1-7}
   (A) Zero & 0.583 & 0.632 & 0.607 & 0.459 & 0.601 &  0.521 \\
   (B) Web & 0.755 & 0.833 & 0.792 & 0.873 & 0.655 & 0.749 \\
   (C) KB & 0.861 & 0.855 & 0.858 & 0.937 & 0.764 & 0.842 \\
   (D) KG & 0.786 & 0.772 & 0.779 & 0.906 & 0.705 &  0.793 \\
   (E) $\mathrm{Fuse}_{\mathrm{C}}$ & 0.925 & \underline{0.969} & \underline{0.946} & \textbf{0.995} & \underline{0.864} & \underline{0.925} \\
   (F) $\mathrm{Fuse}_{\mathrm{S}}$ & \underline{0.931} & \textbf{0.972} & \textbf{0.951} & \underline{0.990} & {0.808} & {0.890} \\
   (G) ENSB & \textbf{0.934} & \underline{0.969} & \textbf{0.951} & \textbf{0.995} & \textbf{0.868} & \textbf{0.927}\\
    \bottomrule
  \end{tabular}
   \caption{Hallucination detection performance with respect to different evidence sources, where Prec is the abbreviation of Precision and F1 represents the F1 score.}
  \label{tab:detection}
  \vspace{-0.2cm}
\end{table}
\subsection{Retrieval Evaluation (RQ1)}
To verify the necessity of performing multi-source evidence fusion, we experimented to evaluate the quality of retrieval evidence by manually checking whether the evidence could lead to the right answer. 

The experimental results are shown in Table \ref{tab:retrieval}, where HR measures the ratio of the golden evidence in an unranked list, while MRR further considers the position of the golden evidence in a ranked list. 
From the results, we find that `Fuse' performs best in all six cases, which fully demonstrates the effectiveness of fusing evidence from diverse evidence. Besides, KB had a significantly higher recall for golden evidence than Web and KG, which explains why KB performed relatively superior in the following\, detection\, and\, correction.
\begin{table*}[t]
  \renewcommand\arraystretch{1.25}
  \tabcolsep=0.135cm
  \footnotesize
  \centering
  \begin{tabular}{l|cccccc|cccccc}
    \toprule
    \multicolumn{1}{c|}{ \multirow{3}{*}[+0.ex]{\centering \bf \makecell[c]{Evidence \\ Sources}}} & 
    \multicolumn{12}{c}{ \multirow{1}{*}[+0.2ex]{\centering \bf Correctors}} \\
    \cline{2-13}
    
    & \multicolumn{6}{c|}{ \multirow{1}{*}[-0.275ex]{\centering \bf Llama3-8B}} & \multicolumn{5}{c}{ \multirow{1}{*}[-0.275ex]{\centering \bf Qwen2-7B}}\\
    \cline{2-13}
    
    & \multicolumn{1}{c}{ \multirow{1}{*}[-0.275ex]{\centering wo/ cor}} & \multicolumn{1}{c}{ \multirow{1}{*}[-0.275ex]{\centering 1st rnd}} & \multicolumn{1}{c}{ \multirow{1}{*}[-0.275ex]{\centering 2nd rnd}} & \multicolumn{1}{c}{ \multirow{1}{*}[-0.275ex]{\centering 3rd rnd}} & \multicolumn{1}{c}{ \multirow{1}{*}[-0.275ex]{\centering 4th rnd}} & \multicolumn{1}{c|}{ \multirow{1}{*}[-0.275ex]{\centering 5th rnd}} & \multicolumn{1}{c}{ \multirow{1}{*}[-0.275ex]{\centering wo/ cor}} & \multicolumn{1}{c}{ \multirow{1}{*}[-0.275ex]{\centering 1st rnd}} & \multicolumn{1}{c}{ \multirow{1}{*}[-0.275ex]{\centering 2nd rnd}} & \multicolumn{1}{c}{ \multirow{1}{*}[-0.275ex]{\centering 3rd rnd}} & \multicolumn{1}{c}{ \multirow{1}{*}[-0.275ex]{\centering 4th rnd}} & \multicolumn{1}{c}{ \multirow{1}{*}[-0.275ex]{\centering 5th rnd}}\\
    \hline
   (A) Web & \cellcolor{gray!13} & 0.701 & 0.868 & 0.925 & 0.943 & 0.943 & \cellcolor{gray!13} & 0.799 & 0.896 & 0.934 & 0.948 & 0.948 \\
   (B) KB & \cellcolor{gray!13} & \underline{0.758} & 0.899 & 0.948 & \underline{0.966} & \underline{0.966} & \cellcolor{gray!13} & 0.831 & 0.909 & 0.936 & 0.950 & 0.950 \\
   (C) KG  & \cellcolor{gray!13}  & 0.733 & 0.904 &  0.945 & 0.961 & 0.961 & \cellcolor{gray!13} & 0.798 & 0.901 & 0.944 & \underline{0.961} & \underline{0.961} \\
   (D) $\mathrm{Fuse}_{\mathrm{C}}$ & \cellcolor{gray!13} & \textbf{0.794} & \underline{0.924} & \underline{0.964} & \textbf{0.979} & \textbf{0.979} & \cellcolor{gray!13} & 0.840 & \underline{0.939} & \underline{0.960} & \textbf{0.973} & \textbf{0.973} \\
   (E) $\mathrm{Fuse}_{\mathrm{S}}$ & \cellcolor{gray!13} & 0.745 & \textbf{0.927} & \textbf{0.970} & \textbf{0.979} & \textbf{0.979}  & \cellcolor{gray!13} & \textbf{0.880} & \textbf{0.940} & \textbf{0.964} & \textbf{0.973} & \textbf{0.973} \\
   (F) RALE & \cellcolor{gray!13} \multirow{-6}{*}[0ex]{\centering 0.072} & 0.720 & 0.880 & 0.927 & 0.941 & 0.941 & \cellcolor{gray!13} \multirow{-6}{*}[0ex]{\centering 0.072} & \underline{0.859} & 0.922 & 0.944 & 0.948 & 0.948 \\
    \bottomrule
  \end{tabular}
  \caption{Hallucination correction performance, where `wo/ cor' mentions no correction, `rnd' is the abbr of round. What is worth mentioning, 1st rnd represents that the hallucinated content has been corrected one round, and so on.}
  \label{tab:correction}
  \vspace{-0.1cm}
\end{table*}
\subsection{Detection Evaluation (RQ2)}
\label{section:detection}
To verify the effectiveness of the fused evidence and the ensemble classifier, we evaluate the hallucination detection performance on different retrieval sources and the ensemble of the\, retrieval\, sources. 
The experimental results are shown in Table \ref{tab:detection}, where `Zero' means no evidence provided, `$\mathrm{Fuse}_{\mathrm{C}}$' fuses evidence via \textbf{C}oncatenation,  `$\mathrm{Fuse}_{\mathrm{S}}$' fuses evidence via \textbf{S}ummarization, `ENSB' denotes the ensemble classifier. 
(A) performs the worst, indicating the necessity of retrieving external knowledge for detection. 
Comparing (C) with (B) and (D), we find that well-organized KB can offer more clean and supportive evidence than Web and more informative evidence than KG. 
Comparing the fused evidence (\textit{i.e.,} $\mathrm{Fuse}_{\mathrm{C}}$ and $\mathrm{Fuse}_{\mathrm{S}}$) to the evidence from a single source (\textit{i.e,} Web, KB, and KG), we observe that the fused evidence considerably improves detection performance, fully demonstrating the effectiveness of multi-source evidence fusion. 
Our ensemble classifier performs the best in most cases (5 out of 6 cases). The results further indicate the necessity of multi-source evidence fusion.
\subsection{Correction Evaluation (RQ3)}
To verify the effectiveness of hallucination correction, we employ the best-performing detector in Section \ref{section:detection} to check the revised answer. Besides, we only experiment on the hallucinated answer because the\, right\, answer\, does not\, need\, correction. 
The experimental results are shown in Table \ref{tab:correction}, where we employ the approval rate as a metric. From the results, we have the following three observations: (1) If no correction, only 7.2\% of hallucination answers can pass the detection, which indicates that the detector can evaluate the performance of the corrector well. (2) Correcting hallucinations with the fused evidence considerably outperforms that with evidence from a single source, showing the effectiveness of evidence fusion. 
(3) During the 5th round of correction, the approval rate no longer increases compared to the 4th round of that, which suggests a moderate number of rounds is enough. (4) Though detection with the rationale $r$ performs worse than that with the fused evidence $E^F$, the context length of the latter is about five times longer than that of the former.
\section{Related Work}
\subsection{Hallucinations in LLMs}
\label{sec:hallu_in_llms}
While LLMs have demonstrated remarkable capabilities across a range of downstream tasks, a significant concern revolves around their propensity to generate hallucinations~\cite{Zhang2023SirensSI,DBLP:conf/ijcnlp/BangCLDSWLJYCDXF23}. 
Hallucinations can be grouped from different viewpoints.
One prevailing perspective broadly categorizes the hallucination into two types: \textit{Factuality Hallucination} and \textit{Faithfulness Hallucination}~\cite{DBLP:journals/corr/abs-2311-05232}. 
In fact, hallucinations frequently occur in NLP tasks~\cite{Hu2024RefCheckerRF} like summarization~\cite{Maynez2020OnFA, Cao2021HallucinatedBF}, machine translation~\cite{Guerreiro2023HallucinationsIL}, dialog systems~\cite{Honovich2021Q2EF, Dziri2022FaithDialAF} and RAG~\cite{Shuster2021RetrievalAR}. 
This work develops a robust hallucination-checking framework to detect and correct factuality hallucinations in LLMs' generated content.
\subsection{Hallucinations Detection}
Recent studies on hallucination detection mainly focus on factuality hallucinations. 
SelfCheckGPT~\cite{Manakul2023SelfCheckGPTZB} leverages the simple idea that if an LLM knows a given concept, sampled responses are likely to contain consistent facts. 
FactScore~\cite{Min2023FActScoreFA} is a new evaluation way that breaks a generation into a series of atomic facts and computes the percentage of atomic facts supported by a reliable knowledge source.
FacTool~\cite{Chern2023FacToolFD} is a tool-augmented framework, which detects factual errors using tools.
RARR~\cite{Gao2022RARRRA} proposes an intuitive approach by directly prompting LLMs to generate queries, retrieve evidence, and verify actuality. MIND \cite{DBLP:journals/corr/abs-2403-06448} further leverages the internal states of LLMs for real-time detection. 
Despite their effectiveness, these methods generally acquire evidence in a single way, which may fall into the absence of key evidence.
\subsection{Post-hoc editing for factuality}
Recent studies have gone beyond detecting hallucinations to correcting a piece of text to be factually consistent with a set of evidence via post-hoc editing~\cite{Shah2019AutomaticFS,Thorne2020EvidencebasedFE,Balachandran2022CorrectingDF,cao-etal-2020-factual,iso-etal-2020-fact,Gao2022RARRRA,LoganIV2021FRUITFR,Schick2022PEERAC}.
Specifically, FRUIT~\cite{LoganIV2021FRUITFR} and PEER~\cite{Schick2022PEERAC} both implement an editor fine-tuned on Wikipedia edit history to update outdated information and collaborative writing, respectively.
EFEC~\cite{Thorne2020EvidencebasedFE} also implements a full retrieval-and-correct workflow trained on Wikipedia passages ~\cite{Thorne2018FEVERAL}. 
RARR~\cite{Gao2022RARRRA} further considers minimal editing. Albeit studied for ages, very limited works exist in combining multi-round correction with the preservation constraint. 
\section{Conclusion}
This work presents \textsc{Medico}, an innovative hallucination-checking system, which assists users in detecting and correcting factual errors in LLMs' generated content with multi-source evidence fusion. 
To the best of our knowledge, \textsc{Medico} is the first hallucination detection and correction framework that leverages multi-source evidence fusion, provides the rationale behind the decision, as well as revises the incorrect generated content. Last but not least, \textsc{Medico} can not only be used as a tool to help users detect and correct hallucinations in response, but also serve as a security plug-in that automatically checks\, LLMs' replies\, in\, real-time.
\section*{Limitations}
Despite our innovations and improvements, we must acknowledge certain limitations in our work:
\begin{itemize}[leftmargin=*]
    \item \textbf{Noisy Issue.} During the multi-source evidence fusion stage, \textsc{Medico} retrieves evidence from diverse sources, which inevitably brings lots of noise information. Though we have reranked the evidence set, these noises can still slip through the net, which may exercise a negative influence on the following detection and correction. This is the aspect that needs to be improved in the future.
    \item \textbf{Computation Burden.} During the hallucination detection stage, though our proposed ensemble classifier achieves the best performance in most cases, the ensemble classifier uses the LLM likelihood collected from multiple sources as input, considerably increasing the computational burden. Considering the trade-off between computational cost and retrieval accuracy, detecting hallucinations using the fused\, evidence is\, enough.
    \item \textbf{Heuristic Metric.} During the hallucination correction stage, we measure the preservation score based on the character-level Levenshtein edit distance. This metric mechanically measures preservation and may underestimate preservation, as it measures preservation based on characters rather than semantics. Currently, preservation evaluating metrics in the field of LLMs remains an open problem that still\, requires\, further\, investigation.
\end{itemize}
\section*{Ethical Consideration}
Throughout this work, we develop and evaluate our \textsc{Medico} system using an open-source dataset (HaluEval), and two representative open-source LLMs (Llama3-8B and Qwen2-7B), to ensure transparency and integrity in our work. 
One potential risk associated with our work is that \textsc{Medico} supports users to customize retrieval sources by uploading files, which may have data privacy concerns. This is also an essential challenge in the field of LLMs \cite{sun2024trustllm,DBLP:journals/corr/abs-2308-05374}. Therefore, we recommend that users can choose to upload open-access files, rather than\, private\, files.
\section*{Acknowledgments}
This work is jointly supported by grants: National Natural Science Foundation of China (No. 62376067), National Natural Science Foundation of China (No. 62406088), and Guangdong Basic and Applied Basic Research Foundation (2023A1515110078). We sincerely thank all the anonymous reviewers for the detailed and careful reviews as well as valuable suggestions, whose help has further improved our work significantly.
\bibliography{custom}
\bibliographystyle{acl_natbib}

\clearpage
\appendix

\section{Case Study}
We provide some cases to present the procedure of the detection and correction: (1) Table \ref{tab:multi_round} shows the corrector fails to correct the hallucinated content and is not approved by the detector, in the 1st round. Hence, the 2nd round of correction is made and the hallucination content is successfully corrected. 
(2) Table \ref{tab:multi_preser} shows, that in the 1st round, the corrector successfully corrects the hallucinated content but inserts much unnecessary information, which triggers the filtering. Hence, the corrector continues to make corrections until the preservation score $\mathrm{Prev}(o,o^{\prime})$ is greater than or equal to the threshold $\delta$. 
(3) As shown in Table \ref{tab:multi_round_preser}, in the 1st round, the corrector fails to correct the hallucinated content and also inserts much unnecessary information. Hence, the corrector continues to make corrections until the hallucinated content is successfully corrected and the preservation score $\mathrm{Prev}(o,o^{\prime})$ is greater than\, or\, equal\, to\, the\, threshold\, $\delta$,\, simultaneously.
\begin{table}[t]
    \centering
    \small
    \renewcommand{\arraystretch}{1.2} 
    \begin{tabular}{p{0.965\linewidth}} 
        \toprule
        \textbf{Question:} What year did the German composer whose compositions are in The Individualism of Gil Evans die? \\
        \textbf{Right answer:} \textbf{\textcolor[RGB]{0,160,0}{1950}} \\
        \textbf{Hallucinated answer:} Kurt Weill passed away in \textbf{\textcolor[RGB]{173,14,14}{1955}}. \\
        \midrule 
        \textbf{1st round:} Kurt Weill passed away in \textbf{\textcolor[RGB]{173,14,14}{1955}}. \\
        \textbf{Detection:} \textbf{\textcolor[RGB]{173,14,14}{\ding{56}}}  \\
        \textbf{Preservation:} \textbf{\textcolor[RGB]{0,160,0}{\ding{52}}} \\
        \midrule 
        \textbf{2nd round:} Kurt Weill passed away in \textbf{\textcolor[RGB]{0,160,0}{1950}}. \\
        \textbf{Detection:} \textbf{\textcolor[RGB]{0,160,0}{\ding{52}}}  \\
        \textbf{Preservation:} \textbf{\textcolor[RGB]{0,160,0}{\ding{52}}} \\
        \bottomrule
    \end{tabular}
    \caption{A multi-turn correction example from HaluEval, where the right answer and hallucinated answer are marked in \textbf{\textcolor[RGB]{0,160,0}{green}} and \textbf{\textcolor[RGB]{173,14,14}{red}}, respectively.}
    \label{tab:multi_round}
    \vspace{-0.4cm}
\end{table}
\begin{table}[t]
    \centering
    \small
    \renewcommand{\arraystretch}{1.4} 
    \begin{tabular}{p{0.965\linewidth}} 
        \toprule
        \textbf{Question:} What is the stage name of the young female actress who starred in the 2008 American drama Gran Torino directed and produced by Clint Eastwood? \\
        \textbf{Right answer:} \textbf{\textcolor[RGB]{0,160,0}{Ahney Her}} \\
        \textbf{Hallucinated answer:} The actress who starred in the 2008 movie directed by Clint Eastwood and co-starred Christopher Carley and Bee Vang is \textbf{\textcolor[RGB]{173,14,14}{Whitney Cua Her}}. \\
        \midrule 
        \textbf{1st round:} The actress who starred in the 2008 movie directed by Clint Eastwood \hl{who also starred in the film} and co-starred Christopher Carley and Bee Vang is \textbf{\textcolor[RGB]{0,160,0}{Ahney Her}}, \hl{better known by her stage name Ahney Her, is an American actress}. \\
        \textbf{Detection:} \textbf{\textcolor[RGB]{0,160,0}{\ding{52}}}  \\
        \textbf{Preservation:} \textbf{\textcolor[RGB]{173,14,14}{\ding{56}}} \\
        \midrule 
        \textbf{2nd round:} The actress who starred in the 2008 movie directed by Clint Eastwood and co-starred Christopher Carley and Bee Vang is \textbf{\textcolor[RGB]{0,160,0}{Ahney Her}}, \hl{better known by her stage name Ahney Her, is an American actress}.  \\
        \textbf{Detection:} \textbf{\textcolor[RGB]{0,160,0}{\ding{52}}}  \\
        \textbf{Preservation:} \textbf{\textcolor[RGB]{173,14,14}{\ding{56}}} \\
        \midrule 
        \textbf{3rd round:} The actress who starred in the 2008 movie directed by Clint Eastwood and co-starred Christopher Carley and Bee Vang is \textbf{\textcolor[RGB]{0,160,0}{Ahney Her}}.  \\
        \textbf{Detection:} \textbf{\textcolor[RGB]{0,160,0}{\ding{52}}}  \\
        \textbf{Preservation:} \textbf{\textcolor[RGB]{0,160,0}{\ding{52}}} \\
        \bottomrule
    \end{tabular}
    \caption{An example showing that the filtering gets triggered as the edit distance is too large, where the unnecessary information inserted by the corrector is \hl{highlighted}.}
    \label{tab:multi_preser}
\end{table}
\begin{table}[t]
    \centering
    \small
    \renewcommand{\arraystretch}{1.4} 
    \begin{tabular}{p{0.965\linewidth}} 
        \toprule
        \textbf{Question:} Which American restaurant chain and international franchise founded in 1958 that Baiada Poultry is a provider of? \\
        \textbf{Right answer:} \textbf{\textcolor[RGB]{0,160,0}{Pizza Hut}} \\
        \textbf{Hallucinated answer:} Baiada Poultry is a provider of \textbf{\textcolor[RGB]{173,14,14}{Subway}}. \\
        \midrule 
        \textbf{1st round:} Baiada Poultry is a provider of \textbf{\textcolor[RGB]{173,14,14}{Subway}}, \hl{which is an American restaurant chain and international franchise founded in 1958}. \\
        \textbf{Detection:} \textbf{\textcolor[RGB]{173,14,14}{\ding{56}}}  \\
        \textbf{Preservation:} \textbf{\textcolor[RGB]{173,14,14}{\ding{56}}} \\
        \midrule 
        \textbf{2nd round:} Baiada Poultry is a provider of \textbf{\textcolor[RGB]{0,160,0}{Pizza Hut}}, \hl{which is an American restaurant chain and international franchise founded in 1958}. \\
        \textbf{Detection:} \textbf{\textcolor[RGB]{0,160,0}{\ding{52}}}  \\
        \textbf{Preservation:} \textbf{\textcolor[RGB]{173,14,14}{\ding{56}}} \\
        \midrule 
        \textbf{3rd round:} Baiada Poultry is a provider of \textbf{\textcolor[RGB]{0,160,0}{Pizza Hut}}. \\
        \textbf{Detection:} \textbf{\textcolor[RGB]{0,160,0}{\ding{52}}}  \\
        \textbf{Preservation:} \textbf{\textcolor[RGB]{0,160,0}{\ding{52}}} \\
        \bottomrule
    \end{tabular}
    \caption{An example showing that the multi-turn correction is conducted and the edit distance filtering is triggered.}
    \label{tab:multi_round_preser}
\end{table}
\section{Workflow of \textsc{Medico}} 
Algorithm \ref{alg:medico} demonstrates the working flow of the proposed \textsc{Medico} framework. It can be divided into three stages: \textbf{(\uppercase\expandafter{\romannumeral1})} Multi-source Evidence Fusion, \textbf{(\uppercase\expandafter{\romannumeral2})} Hallucination Detection with Evidence, and \textbf{(\uppercase\expandafter{\romannumeral3})} Hallucination Correction with Rationale. In brief, during the stage \uppercase\expandafter{\romannumeral1}, \textsc{Medico} retrieves evidence from diverse sources, then combines and fuses them to get the fused evidence. 
During the stage \uppercase\expandafter{\romannumeral2}, \textsc{Medico} identify hallucinations using the fused evidence or the ensemble of evidence and provide the rationale behind such a decision. During the stage \uppercase\expandafter{\romannumeral3}, \textsc{Medico} performs multi-round corrections until the pre-defined threshold is reached or the detection is approved, where the corrected output $o^\prime$ with lower preservation will\, be\, rejected.
\begin{algorithm*}[!h] 
    \renewcommand{\arraystretch}{1.75} 
    \caption{The Workflow of \textsc{Medico}}
    \label{alg:medico}
    \renewcommand{\algorithmicrequire}{\textbf{Input:}}
    \renewcommand{\algorithmicensure}{\textbf{Output:}}
    \begin{algorithmic}[1]
    
        \REQUIRE User query $q$, the generated content $o$, the hallucination detector $\mathcal{M}_d$ and corrector $\mathcal{M}_c$, the minimum preservation threshold $\delta$.
        \ENSURE The veracity label $v$, the rationale $r$, and the corrected content $o^{\prime}$.
        \STATE Launch the search engine (Web) interface, the knowledge base (KB), and the knowledge graph (KG).
        \STATE  \textbf{\# Step \uppercase\expandafter{\romannumeral1}: Multi-source Evidence Fusion}   
        \STATE Search the $n$ most relevant snippets ${E}^S = \{e^s_1, e^s_2, ..., e^s_{n}\}$ from the Web.
        \STATE Retrieve the $m$ most relevant chunks ${E}^B = \{e^b_1, e^b_2, ..., e^b_{m}\}$ from the KB.
        \STATE Recall the k most relevant linearized triplets ${E}^G = \{e^g_1, e^g_2, ..., e^g_{k}\}$ for the KG.
        \IF{Customized retrieval source provided by users}
        \item Retrieve the $j$ most relevant chunks ${E}^U = \{e^u_1, e^u_2, ..., e^u_{j}\}$ from the UF.
        \ENDIF
        \STATE Get the combined evidence set $E=\{e_1, e_2, ..., e_{n+m+k+j}\}$ with Eq.~(\ref{equ:combine}).
        \STATE Rerank the combined evidence set and get the newly ordered evidence set $\tilde{E} = \{\tilde{e}_1, \tilde{e}_2, ..., \tilde{e}_{l}\}$ with Eq.~(\ref{equ:rerank}).
        \STATE Fuse the newly ordered evidence set and get the fused evidence ${E}^{F}$ with Eq.~(\ref{equ:fuse}).
        \STATE  \textbf{\# Step \uppercase\expandafter{\romannumeral2}: Hallucination Detection with Evidence}   
        \IF{Training classifier}
        \item Compute the LLM likelihood $P = \{p^S, p^B, p^G, p^U, p^F\}$ with Eq.~(\ref{equ:likehood}).
        \item Train a binary classifier (Logistic Regression \cite{DBLP:books/wi/HosmerL00}) using the collected LLM likelihood $P$ with Eq.~(\ref{equ:bce}).
        \item Use the trained classifier to check whether the generated content $o$ has factual errors and output the veracity label $v$.
        \ELSE
            \item Prompt $\mathcal{M}_d$ to check whether the generated content $o$ conflicts with the fused evidence ${E}^{F}$ and output the veracity label $v$.
        \ENDIF
        \STATE Prompt $\mathcal{M}_d$ to generate the corresponding rationale behind such a decision.
        \STATE \textbf{\# Step \uppercase\expandafter{\romannumeral3}: Hallucination Correction with Rationale} 
        \IF{The veracity label $v$ is False}
        \FOR{each $i \in [1, 5]$} 
        \item Identify the hallucinated spans that need to be edited using $\mathcal{M}_c$.
        \item Prompt $\mathcal{M}_c$ to revise these spans separately and output the corrected content $o^{\prime}$.
        \item Prompt $\mathcal{M}_d$ to check whether $o^{\prime}$ has factual errors and output the veracity label $v^{\prime}$.
        \IF{The veracity label $v^{\prime}$ is False}
            \item Continue;
        \ENDIF
        \item Measure the preservation score between $o$ and $o^{\prime}$ with Eq.~(\ref{equ:prev}).
        \IF{The preservation score $\mathrm{Prev}(o,o^{\prime})$ is greater than $\delta$}
            \item Break;
        \ENDIF
        \ENDFOR
        \ELSE
            \item Assign $o$ to $o^{\prime}$.
        \ENDIF
        \RETURN $v$, $r$, $o^{\prime}$
    \end{algorithmic}
\end{algorithm*}

\end{document}